\documentclass{article}
\usepackage{spconf,amsmath,graphicx}

\usepackage{enumitem}
\setlist{nosep, leftmargin=14pt}

\usepackage{mwe} % to get dummy images
\usepackage{cite}
\usepackage{amsmath,amssymb,amsfonts}
\usepackage{algorithmic}
\usepackage{graphicx}
\usepackage{textcomp}
\usepackage{xcolor}

\usepackage{url}

\title{Language Guided Domain Generalized Medical Image Segmentation}
\name{Shahina Kunhimon $^{1}$ , Muzammal Naseer $^1$, Salman Khan $^1$, Fahad Shahbaz Khan $^{1,2}$}
\address{$^{1}$ Mohamed bin Zayed University of Artificial Intelligence - UAE,  \\
 $^{2}$ Linkoping University - Sweden}
\begin{document}
%\ninept
%
\maketitle
\begin{abstract}
Single source domain generalization (SDG) 
holds promise for more reliable and consistent image segmentation across real-world clinical settings particularly in the medical domain, where data privacy and acquisition cost constraints often limit the availability of diverse datasets. Depending solely on visual features hampers the model's capacity to adapt effectively to various domains, primarily because of the presence of spurious correlations and domain-specific characteristics embedded within the image features. 
 Incorporating text features alongside visual features is a potential solution to enhance the model's understanding of the data, as it goes beyond pixel-level information to provide valuable context. Textual cues describing the anatomical structures, their appearances, and variations across various imaging modalities can guide the model in domain adaptation, ultimately contributing to more robust and consistent segmentation. In this paper, we propose an approach that explicitly leverages textual information by incorporating a contrastive learning mechanism guided by the text encoder features to learn a more robust feature representation.  We assess the effectiveness of our text-guided contrastive feature alignment technique in various scenarios, including cross-modality, cross-sequence, and cross-site settings for different segmentation tasks. Our approach achieves favorable performance against existing methods in literature. Our code and model weights are available at  \url{https://github.com/ShahinaKK/LG_SDG.git}.
\end{abstract}

\section{Introduction}
\label{sec:intro}
The generalizability of deep learning-based medical segmentation models often gets compromised by the domain shift between training and test datasets. The discrepancies in  the data acquisition process such as imaging modalities, equipment characteristics, and scanning protocols are the main contributing factors for this distribution shift which is considered as a primary hurdle in their clinical deployment \cite{guan2021domain}.
%of segmentation  models .
Unsupervised Domain Adaptation (UDA) \cite{ganin2015unsupervised} and Multi-Source Domain Generalization (MSDG) \cite{muandet2013domain} approaches, designed to mitigate acquisition shifts, pose practical challenges. These methods depend on training data from the target or multiple source domains, which can be difficult to procure due to cost and privacy concerns. Single Source Domain Generalization (SDG) is a more practical setting for training networks that can handle unseen domains, using training data from a single source domain. %Existing approaches
 Existing SDG techniques leverage techniques such as data augmentation \cite {ouyang2022causality, su2023rethinking}, feature adaptation \cite{zhou2022generalizable, hu2023devil} to improve the data diversity 
 %and informativeness
 and to enhance the model's generalization capabilities.
%%challenges of single source domain generalization
However, SDG comes with its own set of challenges. 
A segmentation model trained on a single source can overfit limiting its ability to generalize to unseen domains. The overfitting arises due to : (i) Dependence on source domain-specific features such as image intensity and texture. For example, in liver segmentation from CT images, liver appears as a high-intensity structure against dark background, creating strong contrast. On the other hand, in T2-SPIR MRI images, liver exhibits varying signal intensities due to water content differences within the structure.
(ii) Spurious correlations due to the confounding variables present in the training image background. For instance, scans from different hospitals may contain varying background objects unrelated to the region of interest (ROI). However, the model might mistakenly correlate them with ROIs when trained on single source data hindering its performance. % A segmentation model trained on images from single source can overfit, limiting its generalization in the unseen target domain. This overfitting arises mainly due to: (i) Dependence on source domain-specific features, such as image appearance (intensities and textures). For example, when dealing with liver segmentation in CT images, it appears as a high-intensity structure against a dark background, creating strong contrast. On the other hand, in T2-SPIR MRI images, liver exhibits varying signal intensities due to water content differences within the liver structures.
% (ii) Spurious correlations due to the confounding variables present in the training image background. 
 % especially when variations in scanning equipment and protocols introduce substantial shifts in these visual features.
%The limitations associated with visual features can be mitigated by incorporating textual input features

%%motivation
To tackle these challenges associated with relying solely on visual features, we utilize language models to communicate visual cues about the ROIs across different domains \cite{lahoud20223d}. During training on a single domain such as CT, we also provide clear descriptions regarding the ROI or label's characteristics including its intensity and structural attributes across both source and target domains (CT and MRI). This multi-domain knowledge equips the model to generalize its understanding from source to target (CT to MRI) and vice versa \cite{awais2023foundational, thawkar2023xraygpt}. We introduce a contrastive approach that aligns textual features with image features, enabling the model to prioritize clinical context over misleading visual correlations. This facilitates the mapping of specific text features to corresponding visual patterns. Our contributions are as follows:

% cross-modality guidance, and domain-invariant features, reducing sensitivity to domain shifts and enhancing adaptability to unseen domains.
% In this paper, we make the following contributions:
\begin{itemize}
    \item We enhance state-of-the-art SDG methods for medical image segmentation by integrating a text-guided contrastive feature alignment module into the training pipeline. This provides a domain-agnostic perspective, reducing sensitivity to domain shifts and spurious correlations by grounding visual features with textual information.
    
    \item Our proposed approach is complementary and can be seamlessly integrated into any segmentation network without architectural modifications.
    
    \item We evaluate our text-guided contrastive feature alignment method in various challenging scenarios including cross-modality, cross-sequence, and cross-site settings for the segmentation of diverse anatomical structures.
\end{itemize}
\section{Related Work}
 SDG methods strive to create robust segmentation models that perform consistently across various clinical scenarios, even in the face of domain shifts caused by differences in acquisition process.
 % Particularly in medical image analysis, where data privacy and acquisition cost constraints often limit the availability of diverse datasets, SDG emerges as a valuable approach.
% existing works
The existing SDG approaches can be mainly categorized into two: (i) Image level adaptation methods such as \cite{zhang2020generalizing, xu2022adversarial, ouyang2022causality, su2023rethinking} which utilizes data augmentation strategies to improve training data diversity and (ii) Image and feature level adaptation methods like \cite{zhou2022generalizable,hu2023devil} which focuses on adapting the feature level representations as well.
%%Causality-inspired Single-source Domain Generalization for Medical Image Segmentation%%
Recently, a causality-inspired data augmentation approach CSDG was introduced in \cite{ouyang2022causality} to mitigate the impact of spurious correlations through causal intervention.
%%Rethinking Data Augmentation for single source domain generalization in medical image segmentation%%
Another augmentation based contrastive SDA approach called SLAug was proposed in \cite{su2023rethinking} which incorporated class-level information into the augmentation process in order to improve the generalization performance.
% Generalizable Cross-modality Medical Image Segmentation via Style
% Augmentation and Dual Normalization
Among the image and feature level adaptation models, a dual-normalization based method introduced in \cite{zhou2022generalizable} followed a contrastive approach utilizing source-similar or source dissimilar training examples. It performed well under cross-modality settings but its performance is sub-optimal in the cross-site problem setting.  
%%Devil is in channels:Contrastive Single Domain Generalization for medical image segmentation%%
CCSDG approach from \cite{hu2023devil} attempts to improve the generalization performance across data from various centres by utilizing a contrastive feature disentanglement step to filter out the style-sensitive channel representations to enhance the image feature representations. 
\section{Method}
SDG provides a practical solution to mitigate performance degradation caused by distribution shifts in medical image segmentation models. Instead of requiring data from multiple sources or target domains, 
%it focuses on training with a single source dataset to adapt to various unseen target domains.
%In SDG problem setting for segmentation,
%For medical image segmentation in SDG problem setting, 
SDG focuses on training the network on single source dataset $\mathcal{D}^s=\{x, y\}$ where $x$ represents the image and $y$ corresponds to the ground truth, with the objective of generalizing well to unseen target domains.
%$\mathcal{D}^u=\{x^{u}\}.

\begin{figure}[!t]
  \centering
  \includegraphics[width=0.49\textwidth]{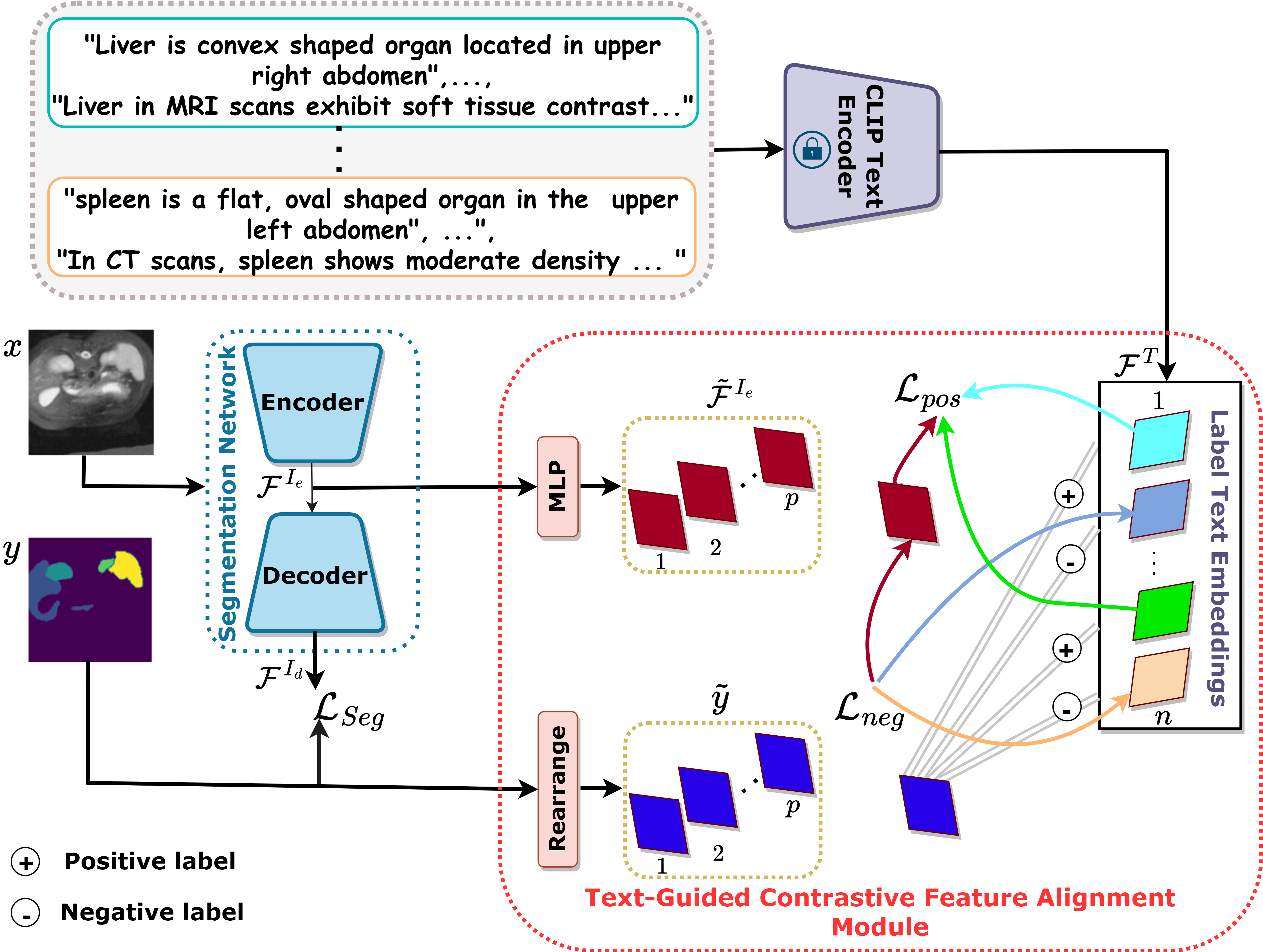}
    \caption{The proposed  training pipeline consists of \textbf{(i) Segmentation network:} an encoder-decoder model \textbf{(ii) CLIP Text Encoder:} which is frozen and takes text descriptions from ChatGPT as input to create label-wise text embeddings and \textbf{(iii) Text-Guided Contrastive Feature Alignment Module:} which enhances the alignment between the image and text encoder representations via our feature-level contrastive loss.} 
  % \caption{The proposed SDG training pipeline for medical image segmentation consists of \textbf{(i) Segmentation network:} which encodes the input image. \textbf{(ii) CLIP Text Encoder:} which is kept frozen and takes the text descriptions from ChatGPT as input to create label-wise text embeddings, and \textbf{(iii) Text-Guided Contrastive Feature Alignment Module:} where the patch-wise contrastive feature alignment between the image and text encoder representations is performed.} 
  \label{fig_main}
\end{figure}

Different from existing approaches, our approach addresses the SDG problem in medical image segmentation by exploiting  well-structured language models 
%pretrained on extensive datasets.
that have already been trained on large-scale datasets. 
We use ChatGPT to generate diverse organ-specific text descriptions for all the possible segmentation classes.
%These descriptions capture various facets of each label to distinguish labels across source and target domains. 
For instance; in the cross-modality (CT/MRI) experiments on Abdomen dataset, we pass the following query to ChatGPT:
\noindent \emph{``Describe the appearance, texture, size, shape, intensity and other characteristics to distinguish the liver, right kidney, left kidney, and spleen from each other in CT and MRI".}
\underline{Sample Response:} 
\noindent \emph{``The liver in CT images appears as a high-intensity structure with uniform texture whereas in MRI, the liver exhibits varying signal intensities."}
These descriptions capture various facets of labels to distinguish them across source and target domains.
It's important to note that not all image crops necessarily contain all segmentation classes and there can be instances where anatomical structures overlap or share visual traits. To address these class-level ambiguities, we introduce a text-guided contrastive feature alignment (TGCFA) module. This module extracts class-level information from ground truth, to guide image representation learning so that the visual features learned lie closer to the text embeddings of the positive classes.
%%%%
% We  formulate a supervised contrastive objective that uses feature-level label information and static text feature representations to guide image representation learning so that the features learned lie closer to the text embeddings of the same semantic class.
%%%%
The proposed method incorporates a text encoder and TGCFA module complementing the baseline image segmentation network as shown in  Fig.\ref{fig_main}.
% Also, as segmentation task highly benefits from the pixel-level labels, we utilize the ground truth masks for class-level region details about location and boundaries to enhance image and text feature alignment.
 \textbf{Baseline Segmentation Network:} We employ a U-shaped encoder-decoder network
  $\mathcal{F}^{I}$ as the segmentation backbone. For each image $x$, the segmentation network outputs the encoded feature vector $\mathcal{F}^{I_{e}} \in \mathbb{R}^{p\times z}$ and segmentation prediction $\mathcal{F}^{I_{d}}$, where $p$ denotes the number of features and $z$ represents the embedding dimension of each feature. The final segmentation mask ($\mathcal{F}^{I_{d}}$) is compared with ground truth $y$ to optimize the segmentation objective ($\mathcal{L}_{Seg}$) such as cross-entropy or dice loss. The encoder representation ($\mathcal{F}^{I_{e}}$) is then fed to our TGCFA module for text-guided contrastive feature alignment.
%%%%dataset summary-Table%%%%%%%
\begin{table*}[t]
\scriptsize
\centering
\caption{Details of the Cross-Domain Segmentation datasets and the source-target splits used for evaluation
%Three datasets from different cross domain settings were used for evaluation.
}
\begin{center}
\centering\setlength{\tabcolsep}{8pt}
\begin{tabular}{|c|c|c|c|c|c|c}
\hline
%\textbf{Table}&\multicolumn{3}{|c|}{\textbf{Table Column Head}} \\
%\cline{2-4} 
% \textbf{Head} & \textbf{\textit{Table column subhead}}& \textbf{\textit{Subhead}}& \textbf{\textit{Subhead}} \\
\textbf{Name} &\textbf{Dataset} & \textbf{Labels} & \textbf{Split} & \textbf{Domains} &\textbf{ No. of scans}\\
\hline
Abdomen cross-modality & SABSCT \cite{landman2015miccai} & Liver, Right kidney,  & Source/Target & CT &30\\
                         &  CHAOS  \cite{kavur2021chaos} & Left kidney, Spleen  & Target/Source & MRI &20\\
\hline
Cardiac cross-sequence & MS-CMR \cite{zhuang2022cardiac} & Myocardium, Left ventricle, & Source/Target & balanced steady-state free precession(bSSFP)   & 45\\
                         &   & Right ventricle  & Target/Source & late gadolinium enhanced(LGE)&45\\
\hline
Fundus cross-site & RIGA+ \cite{hu2022domain} & Optic Disc, & 1 source  & BinRushed/Magarabia & 195, 95\\
                         &   & Optic Cup  & 3 target &  BASE 1, BASE 2, BASE 3 & 173, 148, 133 \\
\hline
\end{tabular}
\label{tab1}
\end{center}
  \vspace{-3 mm}
\end{table*}
%%%%%%%%%%SLAug Results%%%%%%%%%%%%%%%%%%%
\begin{table*}[t]
\scriptsize
\caption{For Cross-Modality Experiments (Left): Our approach consistently improves the organ-wise results and narrows down the gap between SDG and supervised upperbound (trained and tested in target domain). For Cross-Sequence Experiments (Right): Our approach improves the average dice score in both directions. The highest scores in SDG setting are highlighted.  }
\begin{center}
\centering\setlength{\tabcolsep}{9pt}
\begin{tabular}[\textwidth]{|c| c c c c |c|| c c c |c|}
\hline
\textbf{Method}&\multicolumn{5}{c||}{\textbf{ Cross Modality (Abdominal CT-MRI)}} & \multicolumn{4}{c|}{\textbf{Cross-Sequence (Cardiac bSSFP-LGE) }}  \\
\cline{2-10} 
               & Liver& Right Kidney& Left Kidney& Spleen  &  Average & Left Ventricle& Myocardium & Right Ventricle& Average \\
\hline
Supervised & 91.30 & 92.43  & 89.86 & 89.83  &  90.85 & 92.04 & 83.11 & 89.30 & 88.15 \\

SLAug  \cite{su2023rethinking}    &90.08 & 89.23  & 87.54 & 87.67 & 88.63 & \textbf{91.53} & 80.65 & 87.90 & 86.69 \\

\hline
SLAug(Ours)& \textbf{90.52} & \textbf{89.54}  & \textbf{88.21} & \textbf{88.44} & \textbf{89.18} & 91.50 & \textbf{81.10} &  \textbf{88.24} & \textbf{86.95}  \\
\hline
\hline
\textbf{Method}&\multicolumn{5}{c||}{\textbf{Cross Modality (Abdominal MRI-CT)}} & \multicolumn{4}{c|}{\textbf{Cross-Sequence (Cardiac LGE-bSSFP)}}  \\
\cline{2-10} 
               & Liver& Right Kidney& Left Kidney& Spleen& Average  & Left Ventricle& Myocardium & Right Ventricle& Average \\
\hline
Supervised & 98.87 & 92.11  & 91.75 & 88.55 & 92.82 & 91.16 & 82.93 & 90.39 & 88.16 \\

SLAug \cite{su2023rethinking}     &89.26 & 80.98  & 82.05 & 79.93 & 83.05 & \textbf{91.92} & 81.49 & 89.61 & 87.67 \\

\hline
SLAug(Ours)& \textbf{90.34} & \textbf{84.14}  & \textbf{82.15} & \textbf{84.76} & \textbf{85.35} & 91.87 & \textbf{81.93}  & \textbf{89.88}  & \textbf{87.89} \\
\hline
\end{tabular}
\label{ta2}
\end{center}
\vspace{-2 mm}
\end{table*}
%%%%%%%%%%%%%
\noindent\textbf{Text Encoder:} We leverage a pretrained CLIP \cite{radford2021learning} text encoder which remains frozen to map the label descriptions to text feature representation. 
% The text encoder remains frozen and we use chatGPT to create multiple textual descriptions for each label-class. 
Specifically, for an $n$-class segmentation problem, we generate $v$ variations of text descriptions for each of the $n$ labels  as
  $t_r=\{ t_{r_1}, t_{r_2}, ..., t_{r_v} \}_{r=1}^{n}$.
 % These diverse descriptions capture various facets of each label across domains, aiding the model in learning how to distinguish labels across source and target domains. 
These descriptions are tokenized and passed through pretrained CLIP text encoder to get corresponding text embeddings $O_r=\{ O_{r_1}, O_{r_2}, ..., O_{r_v} \}_{r=1}^{n}$. We compute the mean representation for each label to get the text feature vector $\mathcal{F}_{r}^{T} \in \mathbb{R}^{k}$ for label $r$ where $k$ is the dimensionality of text representation i.e, the final text feature representation is given by:
\begin{equation}
\begin{gathered}[b]
    \mathcal{F}^{T} = \{ \mathcal{F}_{1}^{T},\mathcal{F}_{2}^{T}, \hdots \mathcal{F}_{n}^{T} \} %\in  n\times %\mathbb{R}^{k}
    \\
    \text{where} \hspace{1 em}\mathcal{F}_{r}^{T}= \frac{\Sigma(O_{r_1}, O_{r_2}, ..., O_{r_v}) }{v}  \in \mathbb{R}^{k}  
\end{gathered}    
\end{equation}
% Here, $n$ refers to the number of classes and $d$ corresponds to the dimensionality of text representation space.
%\textcolor{red}{add sample text output}\\
%  
\textbf{Text-Guided Contrastive Feature Alignment (TGCFA) module:} 
The text representations $\mathcal{F}^{T}$ and image encoder output $\mathcal{F}^{I_{e}}$ along with the ground truth $y$ are used in the TGCFA module to learn the multi-modal feature alignment objective.
To make the multi-modal (image and text) representations consistent, the visual encoder features  $\mathcal{F}^{I_{e}} \in  \mathbb{R}^{p\times z}$ undergo projection to match the dimension of the text representation  which results in $\tilde{\mathcal{F}}^{I_{e}} \in \mathbb{R}^{p \times k}$. The ground truth $y$ is rearranged to get feature-level masks $\tilde{y}  \in \mathbb{R}^{p \times w} $ followed by the extraction of positive and negative labels for the feature-level alignment.
% so that we can extract the patch-wise class information to facilitate the patch-level feature alignment.
% In our approach,we have formulated a supervised contrastive objective which utilizes the pixel-level semantic label information to guide the representation learning by using semantic text embeddings $\mathcal{F}^{T} \in n \times \mathbb{R}^\mathbf{D} $ 
% %and the patchified segmentation labels $\tilde{y} \in p \times \mathbb{R}^\mathbf{M}$ 
% to anchor the projected image encoder representation $\tilde{\mathcal{F}}^{I} \in p \times \mathbb{R}^\mathbf{D}$ such that the features learnt lie closer to the text embeddings of the positive classes.
\textbf{Feature-level contrastive alignment loss:} Given the positive label set $\mathcal{C}_{j}^{+}$ and negative label set $\mathcal{C}_{j}^{-}$ for each feature-level mask $\tilde{y}_j : j \in [1,p]$, our feature-level contrastive alignment loss $\mathcal{L}_{Align}$ could be represented as
    \begin{equation}\label{eqn_lalign}
    \vspace{-1 mm}
    \mathcal{L}_{Align} = \mathcal{L}_{pos} + \mathcal{L}_{neg}
    \vspace{-1 mm}
    \end{equation}
where we map together the feature level image encoder representation  $\tilde{\mathcal{F}}_{j}^{I_{e}}$ to the text feature $\mathcal{F}_{m}^{T}$ for all the corresponding positive labels $m$ in the corresponding mask.
\begin{equation}\label{eqn_lpos}
    \mathcal{L}_{pos} = \sum_{j=1}^{p} \frac{1}{\left|\mathcal{C}_{j}^{+} \right|}\sum_{m \in \mathcal{C}_{j}^{+} } \max\bigg( 0,1- \bigg[ \frac{\tilde{\mathcal{F}}_{j}^{I_{e}}.\mathcal{F}_{m}^{T}}{\left| \tilde{\mathcal{F}}_{j}^{I_{e}} \right|. \left|\mathcal{F}_{m}^{T} \right| } \bigg] \bigg)
\end{equation}

\noindent Similarly, the possible non-matching label features $\mathcal{F}_{q}^{T}$ 
%for all the negative labels $q$
are pushed away from the visual encoder representations.
\begin{equation}\label{eqn_lneg}
    \mathcal{L}_{neg} = \sum_{j=1}^{p} \frac{1}{\left|\mathcal{C}_{j}^{-} \right|}\sum_{q \in \mathcal{C}_{j}^{-}} \max\bigg( 0,\bigg[ \frac{\tilde{\mathcal{F}}_{j}^{I_{e}}.\mathcal{F}_{q}^{T}}{\left| \tilde{\mathcal{F}}_{j}^{I_{e}} \right|. \left|\mathcal{F}_{q}^{T} \right| } \bigg]-1 \bigg)
\end{equation}

By integrating our feature-level  contrastive alignment loss, the total loss for supervised training becomes:
\begin{equation}\label{eqn_ltotal}
\vspace{-1 mm}
\mathcal{L} = \mathcal{L}_{Seg} + \mathcal{L}_{Align}
\vspace{-1 mm}
\end{equation}

\section{Experiments and Results}
\textbf{Datasets and Experimental setup:}
We evaluate our approach in three cross domain settings: 1) cross-modality abdomen, 2) cross-sequence cardiac and 3) cross-site fundus datasets. Dataset details are summarized in Table \ref{tab1}. \textbf{Training Details:} For cross-modality and cross-sequence experiments, all models are trained on U-Net with Efficient Net-b2 backbone for 2K epochs with inputs size 192×192 and initial learning rate of $3\times 10^{-4}$. For fair comparison with SLAug \cite{su2023rethinking}, we follow the same preprocessing steps and dataset split. For cross-site experiments, we used the same training pipeline and preprocessing steps of CCSDG \cite{hu2023devil} and trained models at 592×592 resolution for 100 epochs with initial learning rate $10^{-2}$. We chose BinRushed and Magrabia as distinct source domains for training and evaluated the model on BASE1, BASE2 and BASE3 target domains. \textbf{Evaluation Metric:} Percentage Dice similarity coefficient \cite{milletari2016v} is used as the evaluation metric to measure the segmentation performance.
\begin{table}[t!]
\vspace{-3 mm}
\scriptsize
\caption{\textbf{Quantitative results (Cross-Site Fundus):} Our approach outperforms supervised training and CSSDG approach. OC refers to Optic Cup and OD refers to Optic Disc.}
\centering
\setlength{\tabcolsep}{3pt} % Reduce inter-column space
\begin{tabular}{|c|p{.65cm} p{.65cm}|p{.65cm} p{.65cm}|p{.65cm} p{.65cm}|p{.65cm} p{.65cm}|}
\hline
\textbf{Method} & \multicolumn{2}{c|}{\textbf{BASE1}} & \multicolumn{2}{c|}{\textbf{BASE2}} & \multicolumn{2}{c|}{\textbf{BASE3}} & \multicolumn{2}{c|}{\textbf{Average}} \\
\cline{2-9}
 & OD & OC & OD & OC & OD & OC & OD & OC \\      
\hline
Supervised & 94.71 &84.07 & 94.84 & 86.32 & 95.40 & 87.34 & 94.98 & 85.91\\
\hline
\hline
\multicolumn{9}{|c|}{\textbf{BinRushed}}\\
\hline
\hline
CCSDG \cite{hu2023devil}& 95.73 & 86.13 & 95.73 & 86.82 & 95.45 &86.77 & 95.64 & 86.57\\
\hline
CCSDG(Ours) & \textbf{ 96.06} & \textbf{86.27} & \textbf{96.03} & \textbf{87.90} & \textbf{95.77} & \textbf{87.07} & \textbf{95.95} & \textbf{87.08}\\
\hline
\hline
\multicolumn{9}{|c|}{\textbf{Magrabia}}\\
\hline
\hline
CCSDG \cite{hu2023devil}& 94.78 & \textbf{84.94} & 95.16 &85.68 & \textbf{95.00} & \textbf{85.98} & 94.98 & 85.53\\
\hline
CCSDG(Ours) &  \textbf{95.05} & 84.55 & \textbf{95.66} & \textbf{88.32} & 94.92 & 85.45 & \textbf{95.21} & \textbf{86.10} \\
\hline
\end{tabular}
\label{ta3}
\end{table}
%%%%%%%%%@@@@%%%%%%%%%%%%
\begin{figure}
\centering
\includegraphics[width=0.43\textwidth]{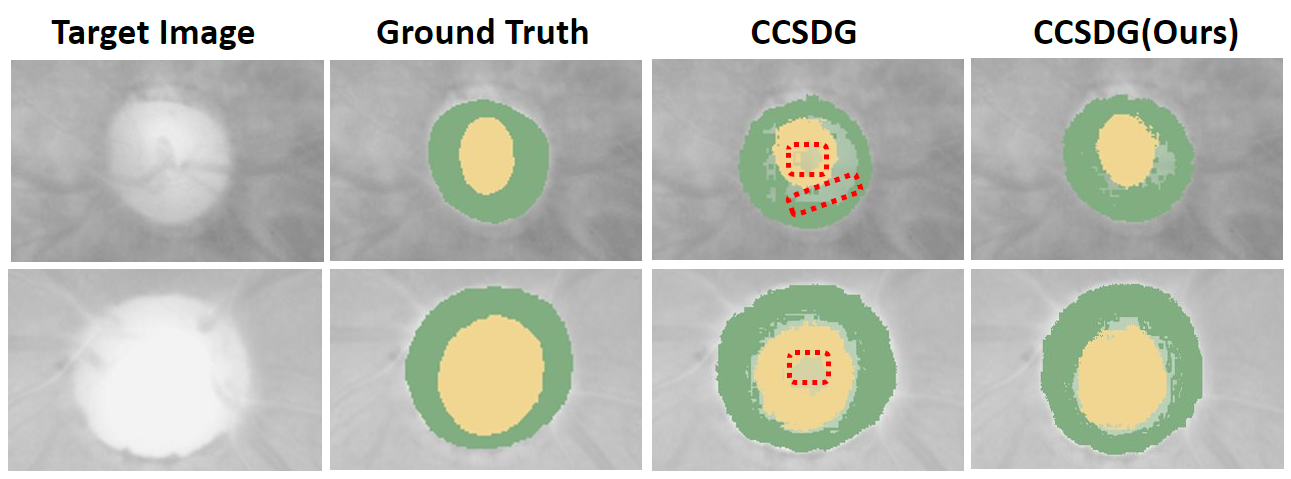}
\vspace{-4 mm}
\caption{\textbf{Qualitative Results (Cross-Site Fundus):} CCSDG struggles to accurately define label boundaries (red dashed box), while our approach enhances boundary definition.}
\label{fig3}
\end{figure}
\textbf{Result Analysis:} Overall, our proposed approach consistently outperforms the baseline methods in all three problem settings. The results of cross-modality Abdomen datasets (Table \ref{ta2}-\emph{left}) indicates that our approach excels, particularly in the MRI-CT generalization task ($2.3 \% \uparrow$) where the performance gap between intra-domain supervised training and SLAug is high. In the cardiac cross-sequence scenario where complexities in resolution are prevalent, our approach manages to enhance the SLAug performance (Table \ref{ta2}- \emph{right}) despite the small domain gap. For cross-site experiments, incorporating our text-guided contrastive feature alignment approach enhances the model's comprehension of organ structures and mitigates the impact of background artifacts, leading to improvements in segmentation results across different centers as depicted in Table \ref{ta3}. 
\textbf{Enhanced Organ Boundary Delineation:} In addition to improvements in the dice score, the qualitative results reveal that our approach enhances the refinement of ROI boundaries and reduces instances of miss classification. This enhancement is crucial for precise organ localization. Visualizations presented in Fig.\ref{fig2} shows that our approach significantly improves the model's capability to accurately locate and delineate organ boundaries in both CT-MRI and MRI-CT settings. Notably, in the context of cardiac cross-sequence experiments, although the dice score improvement is modest, our approach excels in the precise delineation of organ boundaries, as visually depicted in Fig. \ref{fig4}. Qualitative results from cross-site experiments, as shown in Fig.\ref{fig3}, further emphasize that our approach generates accurate segmentation masks by meticulously refining organ boundaries with precision and sharpness.
\begin{figure}[t!]
\begin{minipage}[b]{1.0\linewidth}
  \centering
  \centerline{\includegraphics[width=7.5cm]{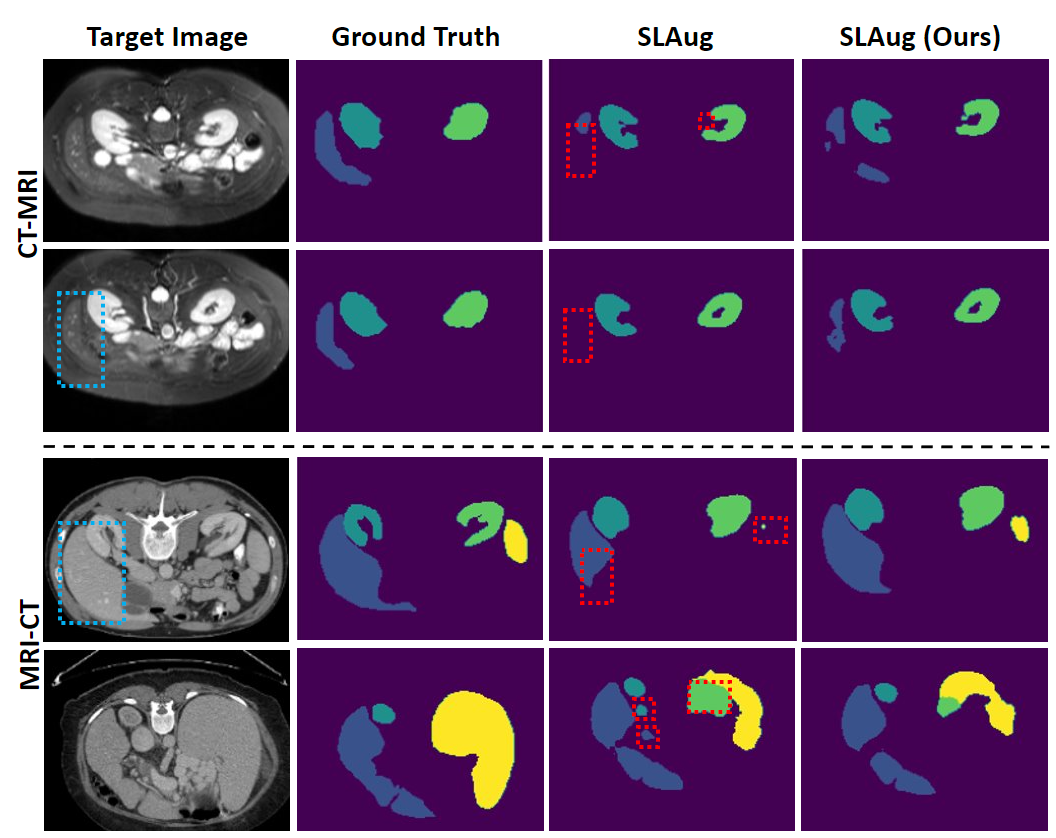}}
   \caption{ \textbf{Qualitative results (Cross-Modality Abdomen):} Domain-specific appearance shift :- Liver \emph{(blue dashed box)} appears dark in MRI \emph{(top)} and bright in CT \emph{(bottom)} images. Our approach enhances the SLAug baseline by reducing miss classification \emph{(red dashed box )} and refining organ boundaries. } 
   %\medskip
   \label{fig2}
 \end{minipage}
 %\vspace{2 mm}
 \begin{minipage}[b]{1.0\linewidth}
   \centering
  \centerline{\includegraphics[width=7.5cm]{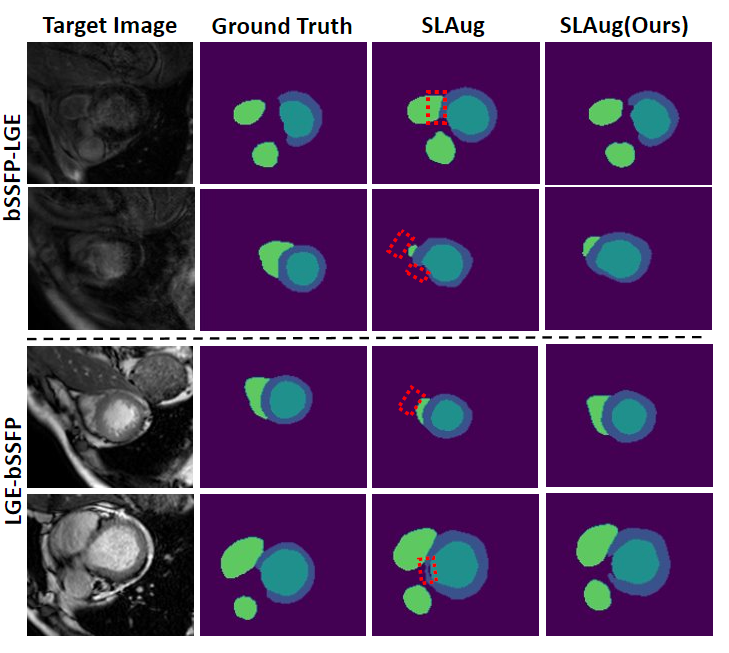}}
  \caption{\textbf{Qualitative results (Cross-Sequence Cardiac):} Our approach outperforms the SLAug baseline in delineating organ boundaries (highlighted by red dashed boxes).} %\medskip
  \label{fig4}
 \end{minipage}
\end{figure}

\section{Conclusion}
This paper introduces an enhancement to single source domain generalization in medical image segmentation by combining textual information with visual features, effectively addressing domain shifts and improving segmentation robustness. The proposed text-guided contrastive feature alignment method demonstrates its efficacy, bringing significant improvements in challenging clinical scenarios, including cross-modality, cross-sequence, and cross-site settings.

\section{Compliance with ethical standards}
\label{sec:ethics}
This study was conducted retrospectively using human subject datasets made available in open access by \cite{landman2015miccai}, \cite{kavur2021chaos}, \cite{zhuang2022cardiac}, and \cite{hu2022domain}. Ethical approval was not required, as confirmed by the licenses attached to the open access data.

\section{Acknowledgments}
\label{sec:acknowledgments}
The authors gratefully acknowledge the scientific support and High Performance Computing (HPC) resources provided by Mohammed bin Zayed University of Articial Intelligence. No funding was received for conducting this study. The authors have no relevant financial or non-financial interests to disclose.

\bibliographystyle{IEEEbib}
\bibliography{main}

\end{document}